\documentclass[10pt,twocolumn,letterpaper]{article}

\usepackage[pagenumbers]{cvpr} 

\usepackage{graphicx}
\usepackage{amsmath}
\usepackage{amssymb}
\usepackage{booktabs}
\usepackage{multirow}

\usepackage[pagebackref,breaklinks,colorlinks]{hyperref}

\usepackage[capitalize]{cleveref}
\crefname{section}{Sec.}{Secs.}
\Crefname{section}{Section}{Sections}
\Crefname{table}{Table}{Tables}
\crefname{table}{Tab.}{Tabs.}

\def\cvprPaperID{5986} 
\def\confName{CVPR}
\def\confYear{2023}

\begin{document}

\title{Conditional Image-to-Video Generation with Latent Flow Diffusion Models}

\newcommand*\samethanks[1][\value{footnote}]{\footnotemark[#1]}
\author{
Haomiao Ni$^1$\thanks{Work done during the internship at NEC Laboratories America.}
\qquad Changhao Shi$^2$\samethanks
\qquad Kai Li$^3$\qquad Sharon X. Huang$^1$\qquad Martin Renqiang Min$^3$ \\
$^1$The Pennsylvania State University, University Park, PA, USA \\ 
$^2$University of California, San Diego, CA, USA \\
$^3$NEC Laboratories America, Princeton, NJ, USA \\
$^1${\tt\small\{hfn5052, suh972\}@psu.edu}
\quad$^2${\tt\small cshi@ucsd.edu}
\quad$^3${\tt\small\{kaili, renqiang\}@nec-labs.com}
}
\maketitle

\begin{abstract}
\vspace{-2mm}
Conditional image-to-video (cI2V) generation aims to synthesize a new plausible video starting from an image (\eg, a person's face) and a condition (\eg, an action class label like smile). The key challenge of the cI2V task lies in the simultaneous generation of realistic spatial appearance and temporal dynamics corresponding to the given image and condition. In this paper, we propose an approach for cI2V using novel latent flow diffusion models (LFDM) that synthesize an optical flow sequence in the latent space based on the given condition to warp the given image. Compared to previous direct-synthesis-based works, our proposed LFDM can better synthesize spatial details and temporal motion by fully utilizing the spatial content of the given image and warping it in the latent space according to the generated temporally-coherent flow. The training of LFDM consists of two separate stages: (1) an unsupervised learning stage to train a latent flow auto-encoder for spatial content generation, including a flow predictor to estimate latent flow between pairs of video frames, and (2) a conditional learning stage to train a 3D-UNet-based diffusion model (DM) for temporal latent flow generation. Unlike previous DMs operating in pixel space or latent feature space that couples spatial and temporal information, the DM in our LFDM only needs to learn a low-dimensional latent flow space for motion generation, thus being more computationally efficient. We conduct comprehensive experiments on multiple datasets, where LFDM consistently outperforms prior arts. Furthermore, we show that LFDM can be easily adapted to new domains by simply finetuning the image decoder. Our code is available at \url{https://github.com/nihaomiao/CVPR23_LFDM}.
\end{abstract}

\begin{figure}
    \centering
    \includegraphics[width=\linewidth]{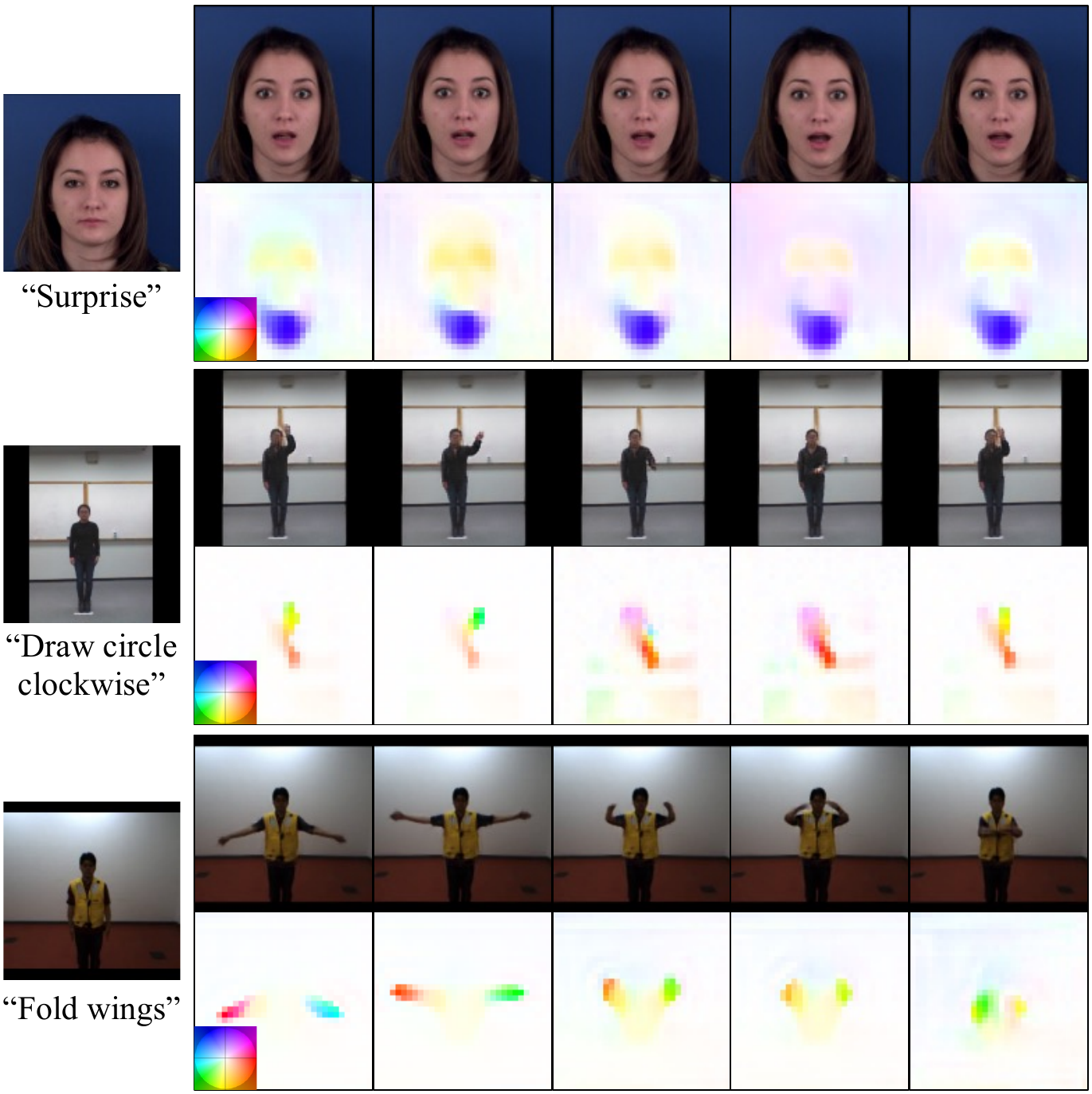}
    \caption{Examples of generated video frames and latent flow sequences using our proposed LFDM. The first column shows the given images $x_0$ and conditions $y$. The latent flow maps are \textit{backward} optical flow to $x_0$ in the \textit{latent} space. We use the color coding scheme in \cite{baker2011database} to visualize flow, where the color indicates the direction and magnitude of the flow.}
    \label{fig:flow}
    \vspace{-2mm}
\end{figure}

\section{Introduction}
Image-to-video (I2V) generation is an appealing topic and has many potential applications, such as artistic creation, entertainment and data augmentation for machine learning \cite{hu2022make}. 
Given a single image $x_0$ and a condition $y$, conditional image-to-video (cI2V) generation aims to synthesize a realistic video with frames $0$ to $k$, $\hat{\mathbf{x}}_0^K=\{x_0, \hat{x}_1, \dots, \hat{x}_K\}$, starting from the given frame $x_0$ and satisfying the condition $y$.
Similar to conditional image synthesis works~\cite{xue2022deep,li2022stylet2i}, most existing cI2V generation methods \cite{dorkenwald2021stochastic,han2022show,he2018probabilistic,hu2022make,kim2019unsupervised,wang2020imaginator} directly synthesize each frame in the whole video based on the given image $x_0$ and condition $y$. However, they often struggle with simultaneously preserving spatial details and keeping temporal coherence in the generated frames.
In this paper, we propose novel latent flow diffusion models, termed {\it LFDM}, which mitigate this issue by synthesizing a latent optical flow sequence conditioned on $y$, to warp the image $x_0$ in the \textit{latent} space for generating new videos (see Fig.~\ref{fig:flow} for an example). Unlike previous direct-synthesis or warp-free based methods, the spatial content of the given image can be consistently reused by our warp-based LFDM through the generated temporally-coherent flow. So LFDM can better preserve subject appearance, ensure motion continuity and also generalize to unseen images, as shown in Sec.~\ref{subsec:results}.

\begin{figure}[t]
    \centering
    \includegraphics[width=\linewidth]{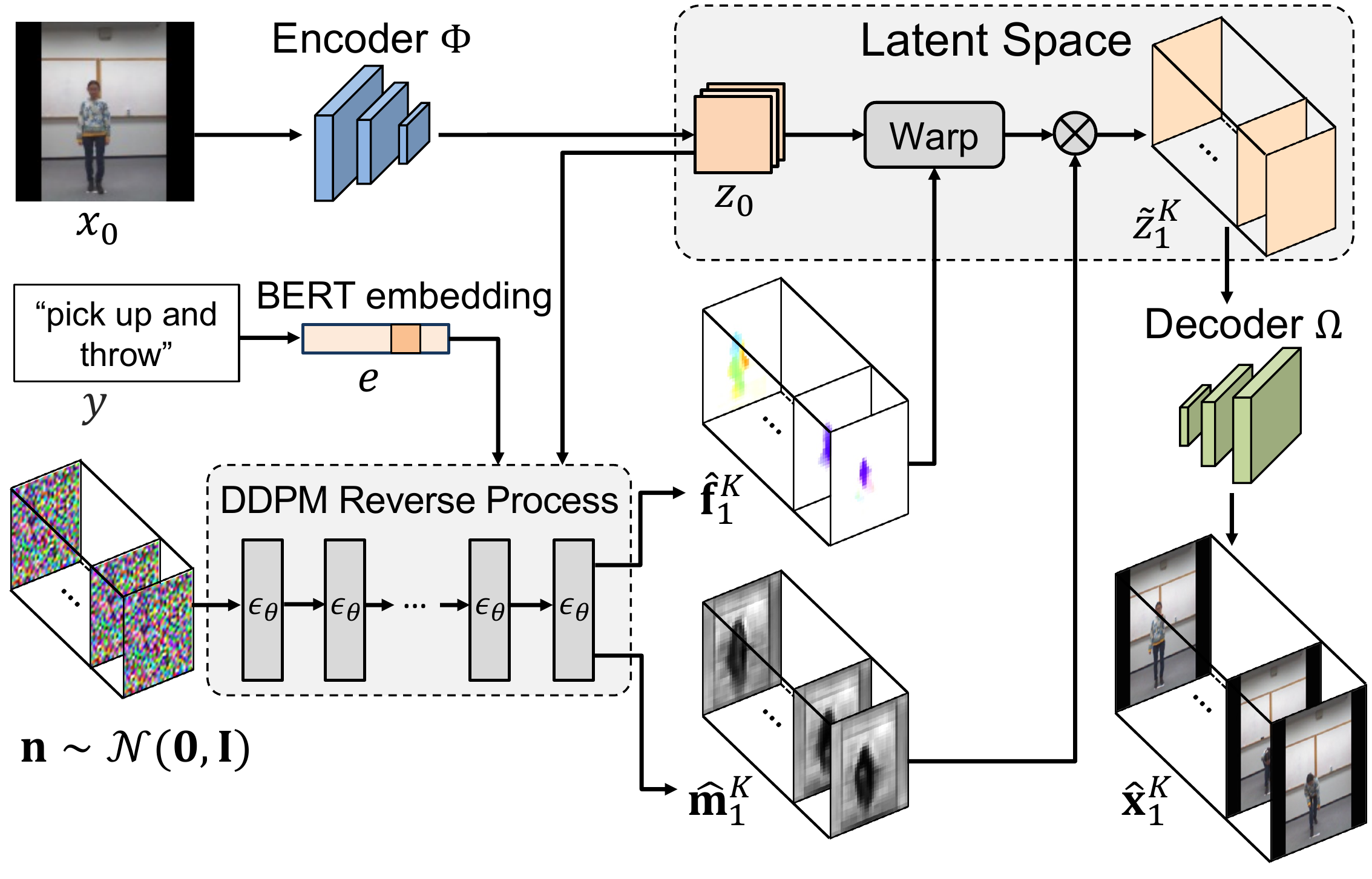}
    \caption{The video generation (\ie, inference) process of LFDM. The generated flow sequence $\hat{\mathbf{f}}_1^K$ and occlusion map sequence $\hat{\mathbf{m}}_1^K$ have the same \textit{spatial} size as image latent map $z_0$. The brighter regions in $\hat{\mathbf{m}}_1^K$ mean those are regions less likely to be occluded.
    }
    \label{fig:testing}
    \vspace{-2mm}
\end{figure}

To disentangle the generation of spatial content and temporal dynamics, the training of LFDM is designed to include two separate stages.
In stage one, inspired by recent motion transfer works \cite{siarohin2019first,Ni_2023_WACV,wang2018video,wang2019few,wiles2018x2face}, a latent flow auto-encoder (LFAE) is trained in an unsupervised fashion. It first estimates the latent optical flow between two frames from the same video, a reference frame and a driving frame. Then the reference frame is warped with predicted flow and LFAE is trained by minimizing the reconstruction loss between this warped frame and the driving frame. 
In stage two, a 3D U-Net \cite{cciccek20163d} based diffusion model (DM) is trained using paired condition $y$ and latent flow sequence extracted from training videos using trained LFAE. Conditioned on $y$ and image $x_0$, the DM learns how to produce temporally coherent latent flow sequences through 3D convolutions.
Different from previous DMs learning in a high-dimensional pixel space \cite{harvey2022flexible,ho2022imagen,ho2022video,singer2022make,voleti2022masked,yang2022diffusion} or latent feature space \cite{rombach2022high} that couples spatial and temporal information, the DM in LFDM operates in a \textit{simple} and \textit{low-dimensional} latent flow space which only describes motion dynamics. So the diffusion generation process can be more computationally efficient.
Benefiting from the decoupled training strategy, LFDM can be easily adapted to new domains. In Sec.~\ref{subsec:results}, 
we show that using the temporal latent flow produced by DM trained in the original domain, LFDM can animate facial images from a new domain and generate better spatial details if the image decoder is finetuned.

During inference, as Fig~\ref{fig:testing} shows, we first adopt the DM trained in stage two to generate a latent flow sequence $\hat{\mathbf{f}}_1^K$ conditioned on $y$ and given image $x_0$. To generate the occluded regions in new frames, the DM also produces an occlusion map sequence $\hat{\mathbf{m}}_1^K$.
Then image $x_0$ is warped with $\hat{\mathbf{f}}_1^K$ and $\hat{\mathbf{m}}_1^K$ in the latent space frame-by-frame to generate the video $\hat{\mathbf{x}}_1^K$. By keeping warping the given image $x_0$ instead of previous synthesized frames, we can avoid artifact accumulation. More details will be introduced in Sec.~\ref{subsec:inference}.

Our contributions are summarized as follows:
\begin{itemize}
    \item We propose novel latent flow diffusion models (LFDM) to achieve image-to-video generation by synthesizing a temporally-coherent flow sequence in the latent space based on the given condition to warp the given image. 
    To the best of our knowledge, we are the first to apply diffusion models to generate latent flow for conditional image-to-video tasks.
    
    \item A novel two-stage training strategy is proposed for LFDM to decouple the generation of spatial content and temporal dynamics, which includes training a latent flow auto-encoder in stage one and a conditional 3D U-Net based diffusion model in stage two. This disentangled training process also enables LFDM to be easily adapted to new domains.
    
    \item We conduct extensive experiments on multiple datasets, including videos of facial expression, human action and gesture, where our proposed LFDM consistently outperforms previous state-of-the-art methods.  

\end{itemize}

\section{Related Work}
\subsection{Image-to-Video Generation}
Conditional video generation aims to synthesize videos guided by user-provided signals.
According to which kind of conditions are provided, conditional video generation can be categorized into text-to-video (T2V) generation \cite{balaji2019conditional,deng2019irc,hong2022cogvideo,li2018video,marwah2017attentive,mittal2017sync,pan2017create,wu2021godiva}, video-to-video (V2V) generation \cite{chan2019everybody,mallya2020world,Ni_2023_WACV,wang2018video,wang2019few,wang2022latent}, and image-to-video (I2V) generation \cite{blattmann2021ipoke,blattmann2021understanding,dorkenwald2021stochastic,hu2022make,mahapatra2022controllable,pan2019video,wang2020imaginator,xiong2018learning,yang2018pose,zhang2020dtvnet}. I2V is also closely related to video prediction from single images \cite{babaeizadeh2017stochastic,li2018flow,kim2019unsupervised}.
Depending on the availability of motion cues, I2V can be further classified into stochastic (\textit{i.e.,} only using given image $x_0$) \cite{babaeizadeh2017stochastic,li2018flow,pan2019video,xiong2018learning,zhang2020dtvnet} and conditional generation \cite{blattmann2021ipoke,blattmann2021understanding,dorkenwald2021stochastic,hu2022make,kim2019unsupervised,mahapatra2022controllable,wang2020imaginator,yang2018pose} (using $x_0$ and given condition $y$). Here we mainly discuss previous conditional image-to-video (cI2V) generation methods. 

Yang \etal \cite{yang2018pose} proposed a pose-guided method by first extracting pose from given image $x_0$ and learning a GAN model to produce pose sequence conditioned on $y$. Then another GAN model was employed to synthesize video frames from the poses.
However, direct pose-to-video synthesis may be difficult to produce fine-grained details. 
Blattmann \etal \cite{blattmann2021ipoke,blattmann2021understanding} proposed interactive I2V models which allowed users to specify the desired motion through manual poking of a pixel. 
Mahapatra \etal \cite{mahapatra2022controllable} achieved cI2V by estimating optical flow maps from motion direction inputs. 
Though these methods do not require users to provide detailed guidance, it is hard for them to generate videos with complex motions. 
Wang \etal \cite{wang2020imaginator} proposed a GAN model ImaGINator, which includes a specially designed spatio-temporal fusion scheme and transposed (1+2)D convolution. Hu \etal \cite{hu2022make} proposed a I2V generator with a motion anchor structure to store appearance-motion aligned representations. However, these direct synthesis methods require modeling both spatial and temporal features, which may complicate the network training. 

\begin{figure*}[t]
    \centering
    \includegraphics[width=\linewidth]{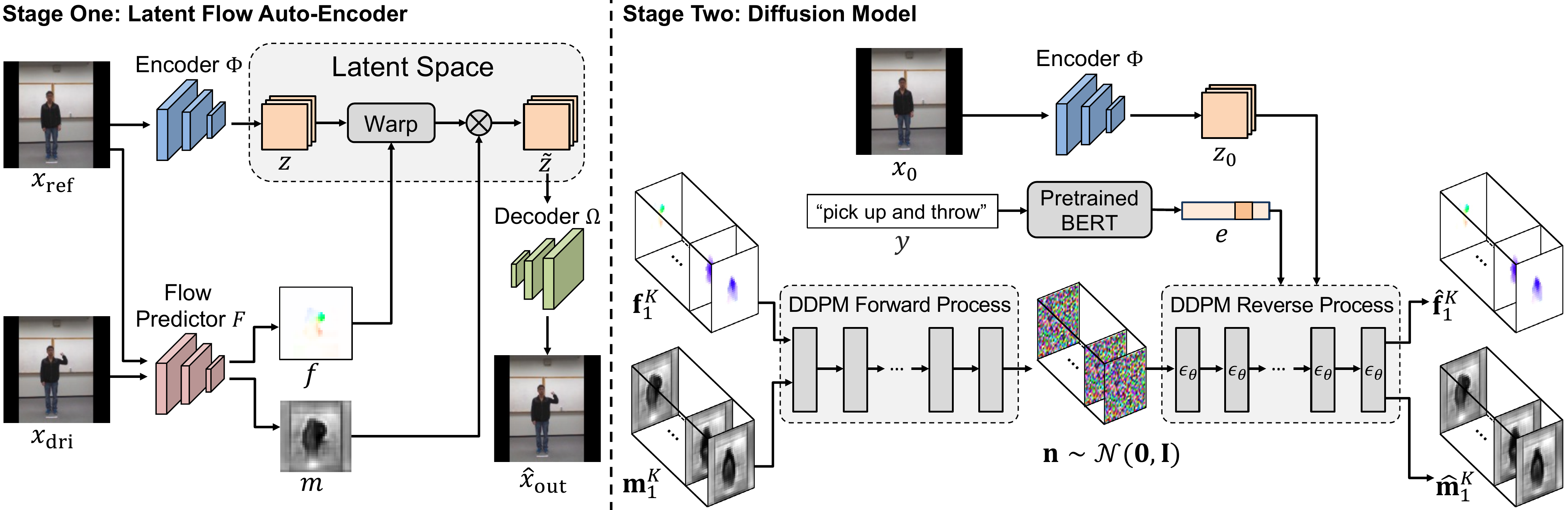}
    \caption{The training framework of LFDM. On the left is stage one for training latent flow auto-encoder while on the right is stage two for training diffusion model. In stage two, the encoder $\Phi$ is the one already trained in stage one, and the latent flow sequence $\mathbf{f}^K_1$ and occlusion map sequence $\mathbf{m}^K_1$ are estimated between $x_0$ and each frame in ground truth video $\mathbf{x}^K_1$ using the trained flow predictor $F$ from stage one.}
    \label{fig:training}
    \vspace{-2mm}
\end{figure*}

\textbf{Differences from previous flow-based I2V works.} 
Several stochastic I2V works \cite{li2018flow,pan2019video,zhang2020dtvnet} designed warping-flow-based models, which generated flow using GAN \cite{goodfellow2020generative} or VAE \cite{kingma2013auto} and only conditioned on given image $x_0$. Our LFDM instead generates flow sequences based on both image $x_0$ and condition $y$ using diffusion models \cite{ho2020denoising}, which have emerged as a new paradigm in generation tasks. Moreover, unlike their separately-trained flow predictor and flow-warped frame generator, the latent flow auto-encoder in LFDM learns these modules together in an unsupervised fashion.
Although the cI2V method in \cite{mahapatra2022controllable} also adopted warping flow to model the motion of fluid elements (water, smoke, fire, \etc), their flow maps can be easily acquired from motion direction inputs. This is because fluid elements can follow the assumption of time-invariant motion field \cite{holynski2021animating}, which does not hold for complex motions such as human movements. 



\subsection{Diffusion Models for Video Generation}
Diffusion models (DMs) \cite{ho2020denoising,sohl2015deep} have very recently found remarkable success in image generation \cite{avrahami2021blended,dhariwal2021diffusion,gu2021vector,ho2021classifier,jiang2022text2human,kim2021diffusionclip,Liu2022CompositionalVG,nichol2021improved,nichol2021glide,preechakul2022diffusion,ramesh2022hierarchical,rombach2022high,saharia2021palette,saharia2022photorealistic}. For example, Rombach \cite{rombach2022high} proposed latent diffusion models (LDM) for image generation by applying DM in the latent space of pretrained auto-encoders.
For video generation, DMs also showed promising performance \cite{harvey2022flexible,ho2022video,ho2022imagen,singer2022make,voleti2022mcvd,yang2022diffusion}.
Ho \textit{et al.} \cite{ho2022video} introduced video diffusion models (VDM) by changing the 2D U-Net \cite{ronneberger2015u} in image DM to be 3D U-Net \cite{cciccek20163d}. In \cite{ho2022imagen}, they further utilized VDM to design a base video generation model and improve it with a sequence of video super-resolution models. Singer \etal \cite{singer2022make} exploited pretrained text-to-image DMs to achieve text-to-video generation without paired text-video data. 
Different from all the above models, LFDM instead applies DM to generate latent flow sequences for conditional image-to-video generation. 

\section{Our Method}
Let $\mathbf{n}\sim\mathcal{N}(\mathbf{0}, \mathbf{I})$ be a Gaussian noise volume with the shape of $K_n \times H_n\times W_n \times C_n $, where $K_n$ $H_n$, $W_n$, and $C_n$ are length, height, width, and channel number, respectively.
Given one starting image $x_0$ and condition $y$, let $\mathbf{x}_0^K=\{x_0, x_1, \dots, x_K\}$ be the real video of condition $y$, the goal of conditional image-to-video (cI2V) generation is to learn a mapping that converts the noise volume $\mathbf{n}$ to a synthesized video, $\hat{\mathbf{x}}_1^K=\{\hat{x}_1, \dots, \hat{x}_K\}$, so that the conditional distribution of $\hat{\mathbf{x}}_1^K$ given $x_0$ and $y$ is identical to the conditional distribution of $\mathbf{x}_1^K$ given $x_0$ and $y$, \ie, $p(\hat{\mathbf{x}}_1^K|x_0, y)=p(\mathbf{x}_1^K|x_0, y)$.
Similar to \cite{kim2019unsupervised,wang2020imaginator,yang2018pose}, we only consider the class label as the input condition $y$.
In this section, we first introduce the background of diffusion models, then explain our proposed two-stage training framework in LFDM and finally illustrate our inference process.

\subsection{Diffusion Models}
Our proposed LFDM is built on denoising diffusion probabilistic models (DDPM) \cite{ho2020denoising,sohl2015deep,song2019generative}. Given a sample from the data distribution $\mathbf{s}_0\sim q(\mathbf{s}_0)$, the \textit{forward} process of DDPM produces a Markov chain $\mathbf{s}_1, \dots, \mathbf{s}_T$ by progressively adding Gaussian noise to $\mathbf{s}_0$ according to a variance schedule $\beta_1, \dots, \beta_T$, that is:
\begin{small}
\begin{equation}
\label{eq:forward}
    q(\mathbf{s}_t|\mathbf{s}_{t-1}) = \mathcal{N}(\mathbf{s}_t; \sqrt{1-\beta_t}\mathbf{s}_{t-1}, \beta_t\mathbf{I})
\enspace,
\end{equation}
\end{small}
where variances $\beta_t$ are held constant.
When $\beta_t$ are small, the posterior $q(\mathbf{s}_{t-1}|\mathbf{s}_{t})$ can be well approximated by diagonal Gaussian \cite{sohl2015deep,nichol2021glide}. Moreover, if the $T$ of the chain is large enough, $\mathbf{s}_T$ can be well approximated by standard Gaussian $\mathcal{N}(\mathbf{0}, \mathbf{I})$. These suggest that the true posterior $q(\mathbf{s}_{t-1}|\mathbf{s}_{t})$ can be estimated by $p_\theta(\mathbf{s}_{t-1}|\mathbf{s}_t)$ defined as:
\begin{small}
\begin{equation}
\label{eq:reverse}
    p_\theta(\mathbf{s}_{t-1}|\mathbf{s}_t)=\mathcal{N}(\mathbf{s}_{t-1}; \mu_\theta(\mathbf{s}_t), \sigma_t^2\mathbf{I})
\enspace,
\end{equation}
\end{small}
where variances $\sigma_t$ are also constants.
The DDPM \textit{reverse} process (also termed \textit{sampling}) then produces samples $\mathbf{s}_0\sim p_\theta(\mathbf{s}_0)$ by starting with Gaussian noise $\mathbf{s}_T\sim \mathcal{N}(\mathbf{0}, \mathbf{I})$ and gradually reducing noise in a Markov chain of $\mathbf{s}_{T-1}, \mathbf{s}_{T-2}, \dots, \mathbf{s}_0$ with learnt $p_\theta(\mathbf{s}_{t-1}|\mathbf{s}_t)$. 
To learn $p_\theta(\mathbf{s}_{t-1}|\mathbf{s}_t)$, Gaussian noise $\epsilon$ is added to $\mathbf{s}_0$ to generate samples $\mathbf{s}_t\sim q(\mathbf{s}_t|\mathbf{s}_0)$, then a model $\epsilon_\theta$ is trained to predict $\epsilon$ using the following mean-squared error loss:
\begin{small}
\begin{equation}
    L=\mathbb{E}_{t\sim \mathcal{U}(1, T), \mathbf{s}_0\sim q(\mathbf{s}_0), \epsilon\sim \mathcal{N}(\mathbf{0}, \mathbf{I})}[||\epsilon-\epsilon_\theta(\mathbf{s}_t, t)||^2]
\enspace,
\end{equation}
\end{small}
where time step $t$ is uniformly sampled from $\{1, \dots, T\}$. Then $\mu_\theta(\mathbf{s}_t)$ in Eq.~\ref{eq:reverse} can be derived from $\epsilon_\theta(\mathbf{s}_t, t)$ to model $p_\theta(\mathbf{s}_{t-1}|\mathbf{s}_t)$ \cite{ho2020denoising}.
Denoising model $\epsilon_\theta$ is usually implemented via a time-conditioned U-Net \cite{ronneberger2015u} with residual blocks \cite{he2016deep} and self-attention layers \cite{vaswani2017attention}. And time step $t$ is specified to $\epsilon_\theta$ by the sinusoidal position embedding \cite{vaswani2017attention}.
For conditional generation, \ie, sampling $\mathbf{s}_0\sim p_\theta(\mathbf{s}_0|y)$, one can simply learn a $y$-conditioned model $\epsilon_\theta(\mathbf{s}_t, t, y)$ \cite{nichol2021glide,rombach2022high}.
Recently, Ho \etal \cite{ho2021classifier} proposed \textit{classifier-free guidance} for conditional generation in DMs. During training, the condition $y$ in $\epsilon_\theta(\mathbf{s}_t, t, y)$ is replaced by a null label $\emptyset$ with a fixed probability. During sampling, the model output is generated as follows:
\begin{small}
\begin{equation}
\label{eq:classifier}
    \hat{\epsilon}_\theta(\mathbf{s}_t, t, y) = \epsilon_\theta(\mathbf{s}_t, t, \emptyset) + g \cdot (\epsilon_\theta(\mathbf{s}_t, t, y)-\epsilon_\theta(\mathbf{s}_t, t, \emptyset))
\enspace,
\end{equation}
\end{small}
where $g$ is the guidance scale.



\subsection{Training}
The overall training process of LFDM is shown in Fig.~\ref{fig:training}, which includes two separate stages to decouple the generation of spatial content and temporal dynamics. In stage one, a latent flow auto-encoder (LFAE) is trained in an unsupervised fashion to estimate latent flow between a pair of video frames, a reference frame and a driving frame, and the reference frame is warped with the latent flow to reconstruct the driving frame. 
In stage two, we train a 3D-UNet-based diffusion model (DM) to produce temporally-coherent latent flow sequence conditioned on image $x_0$ and class $y$. Implementation details will be presented in Sec~\ref{subsec:implement}.
\subsubsection{Stage One: Latent Flow Auto-Encoder}
In stage one, we train a latent flow auto-encoder (LFAE) in an unsupervised manner. As Fig.~\ref{fig:training} shows, LFAE contains three trainable modules: an image encoder $\Phi$ to represent image $x$ as latent map $z$, a flow predictor $F$ to estimate the latent flow $f$ and occlusion map $m$ between video frames, and an image decoder $\Omega$ to decode warped latent map $\Tilde{z}$ as output image $\hat{x}$. During the stage-one training, we first randomly select two frames from the \textit{same} video, a reference frame $x_\text{ref}$ and a driving frame $x_\text{dri}$. Both $x_\text{ref}$ and $x_\text{dri}$ are RGB frames so their size is $H_x \times W_x\times 3$. Encoder $\Phi$ then encodes $x_\text{ref}$ as latent map $z$ with the size of $H_z \times W_z\times C_z$. 
$x_\text{ref}$ and $x_\text{dri}$ are also fed to flow predictor $F$ to estimate the \textit{backward} latent flow $f$ from $x_\text{dri}$ to $x_\text{ref}$, which has the same \textit{spatial} size ($H_z \times W_z\times 2$) as latent map $z$. 
Here flow $f$ has 2 channels because it describes the horizontal and vertical movement between frames.
And we employ \textit{backward} flow because it can be efficiently implemented through a differentiable bilinear sampling operation \cite{jaderberg2015spatial}.
However, only using $f$ to warp latent $z$ may be insufficient to generate the latent map of $x_\text{dri}$ because warping can only use existing appearance information in $z$. When occlusions exist, which are common in those videos containing large motions, LFAE should be able to generate those invisible parts in $z$. 
So similar to \cite{siarohin2019first, wang2018video}, flow predictor $F$ also estimates a latent occlusion map $m$ with the size of $H_z \times W_z\times 1$, as shown in Fig.~\ref{fig:training}. Here $m$ contains values changing from 0 to 1 to indicate the degree of occlusion, where 1 is not occluded and 0 means entirely occluded. The final warped latent map $\Tilde{z}$ can be produced by:
\begin{small}
\begin{equation}
\label{eq:warp}
    \Tilde{z} = m\odot \mathcal{W}(z, f)
\enspace,
\end{equation}
\end{small}
where $\mathcal{W}(\cdot,\cdot)$ is backward warped map and $\odot$ indicates element-wise multiplication.
Decoder $\Omega$ subsequently decodes the visible parts and inpaints the occluded parts of latent $\Tilde{z}$ for generating output image $\hat{x}_\text{out}$, which should be the same as driving frame $x_\text{dri}$. Thus LFAE can be trained with the following reconstruction loss using only \textit{unlabeled} video frames:
\begin{small}
\begin{equation}
\label{eq:fae}
    L_\text{LFAE} = \mathcal{L}_\text{rec}(\hat{x}_\text{out}, x_\text{dri})
\enspace,
\end{equation}
\end{small}
where $\mathcal{L}_\text{rec}$ is the loss measuring the difference between reconstructed frame $\hat{x}_\text{out}$ and ground truth frame $x_\text{dri}$. Here we implement $\mathcal{L}_\text{rec}$ using the perceptual loss \cite{johnson2016perceptual} based on pretrained VGG network \cite{simonyan2014very}.
\subsubsection{Stage Two: Diffusion Model}
In stage two, a 3D-UNet-based diffusion model (DM) is trained to synthesize a temporally-coherent latent flow sequence. The trained image encoder $\Phi$ and flow predictor $F$ from stage one are required in the training of this stage two.
Given an input video $\mathbf{x}_0^K=\{x_0, x_1, \dots, x_K\}$ and its corresponding class condition $y$, we first compute the latent flow sequence from frame 1 to frame $k$,  $\mathbf{f}^K_1=\{f_1, \dots, f_K\}$, and the occlusion map sequence $\mathbf{m}^K_1=\{m_1, \dots, m_K\}$ by applying trained $F$ to estimate $f_i$ and $m_i$ between starting frame $x_0$ and each frame $x_i$, where $i=1, \dots, K$. 
The size of $\mathbf{f}^K_1$ and $\mathbf{m}^K_1$ are $K\times H_z\times W_z\times 2$ and $K\times H_z\times W_z\times 1$, respectively.
We concatenate $\mathbf{f}^K_1$ and $\mathbf{m}^K_1$ along the channel dimension, which produces a $K\times H_z\times W_z\times 3$ volume $\mathbf{s}_0=\mathrm{cat}[\mathbf{f}^K_1, \mathbf{m}^K_1]$. Then $s_0$ is mapped to a standard Gaussian noise volume $\mathbf{n}\sim\mathcal{N}(\mathbf{0}, \mathbf{I})$ by gradually adding 3D Gaussian noise through the DDPM forward process. 
The encoder $\Phi$ further represents starting frame $x_0$ as latent map $z_0$ and pretrained BERT \cite{devlin2018bert} encodes class condition $y$ as text embedding $e$.
Here we choose BERT embedding instead of one-hot encoding to represent $y$ because text embedding can better capture the semantic similarity among different classes. 
Conditioned on $z_0$ and $e$, denoising model $\epsilon_\theta(\mathbf{s}_t, t, z_0, e)$ is trained to predict the added noise $\epsilon$ in $\mathbf{s}_t$ based on a conditional 3D U-Net with the following loss:
\begin{small}
\begin{equation}
\label{eq:fdm}
    L_\text{DM}=\mathbb{E}_{t\sim \mathcal{U}(1, T), \mathbf{s}_0\sim q(\mathbf{s}_0), \epsilon\sim \mathcal{N}(\mathbf{0}, \mathbf{I})}[||\epsilon-\epsilon_\theta(\mathbf{s}_t, t, z_0, e)||^2]
\enspace,
\end{equation}
\end{small}
where time step $t$ is uniformly sampled from $\{1, \dots, T\}$.
$\epsilon_\theta$ is further used in DDPM reverse sampling process to output $\hat{\mathbf{s}}_0=\mathrm{cat}[\hat{\mathbf{f}}_1^K,\hat{\mathbf{m}}_1^K]$ with the size of $K\times H_z\times W_z\times 3$, where $\hat{\mathbf{f}}_1^K=\{\hat{f}_1,\dots,\hat{f}_K\}$ and $\hat{\mathbf{m}}_1^K=\{\hat{m}_1,\dots,\hat{m}_K\}$ are synthesized latent flow and occlusion map sequences.
So the \textit{latent flow space} in our DM is only of the size $K\times H_z\times W_z\times 3$. Its dimensions can be much lower than the RGB pixel space dimensions ($K\times H_x\times W_x\times 3$) used by DMs in \cite{ho2022imagen, ho2022video}, if the spatial size of latent map $H_z\times W_z$ is much smaller than image size $H_x\times W_x$.
Also, the latent flow space only contains motion and shape features. So it can be easier to model than the latent feature space used in \cite{rombach2022high}, which contains other spatial details such as texture and color, as well as all the information in our latent flow space. 
Therefore, the latent flow space in our approach is both \textit{simple} and \textit{low-dimensional}, which helps to ease the generation and reduce computational cost, as shown in Sec.~\ref{subsec:results}.

\subsection{Inference}
\label{subsec:inference}
As Fig.~\ref{fig:testing} shows, 
for a given image $x_0$ and condition $y$, 
the image encoder $\Phi$ encodes $x_0$ as latent map $z_0$ and pretrained BERT represents $y$ as embedding $e$.
Conditioned on $z_0$ and $e$, 
a randomly sampled Gaussian noise volume $\mathbf{n}$ with the size of $K_z\times H_z\times W_z\times 3$ is gradually denoised by $\epsilon_\theta$ through the DDPM reverse sampling process to generate the latent flow sequence $\hat{\mathbf{f}}_1^K$ and occlusion map sequence $\hat{\mathbf{m}}_1^K$.
Then the latent map $z_0$ is warped by each $\hat{f}$ in $\hat{\mathbf{f}}_1^K$ and each $\hat{m}$ in $\hat{\mathbf{m}}_1^K$ according to Eq.~\ref{eq:warp}, producing a new latent map sequence $\Tilde{z}_1^K=\{\Tilde{z}_1, \dots, \Tilde{z}_K\}$. 
Finally, each $\Tilde{z}$ in $\Tilde{z}_1^K$ is further fed to the image decoder $\Omega$ for synthesizing each new frame $\hat{x}$ in output video $\hat{\mathbf{x}}_1^K$. 
Note that flow predictor $F$ is not required during inference.


\begin{figure*}[t]
    \centering
    \includegraphics[width=\linewidth]{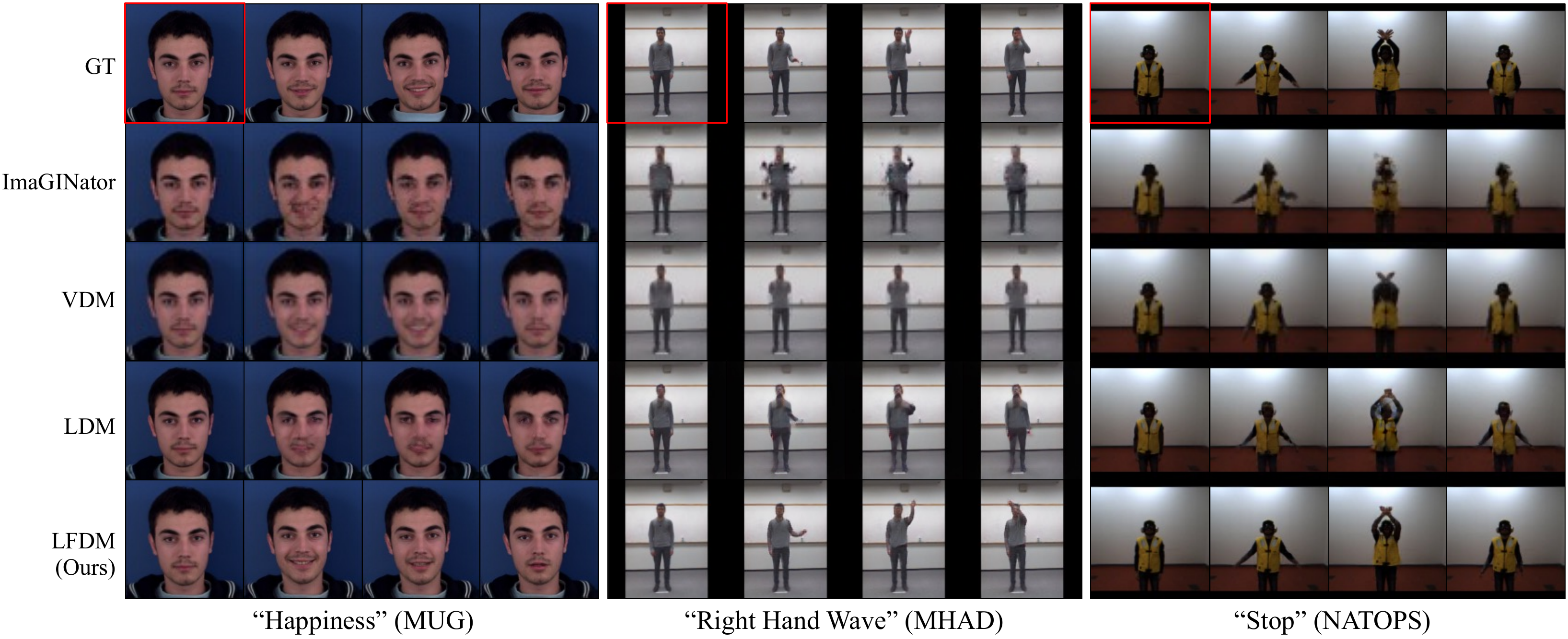}
    \caption{Qualitative comparison among different methods on multiple datasets for cI2V generation. First image frame $x_0$ is highlighted with red box and condition $y$ is shown under each block. To simplify coding, all the models are designed to also generate starting frame $\hat{x}_0$. The video frames of GT (ground truth), LDM and LFDM have $128\times 128$ resolution while results of ImaGINator and VDM are $64\times64$.}
    \label{fig:sota}
    \vspace{-2mm}
\end{figure*}

\begin{table*}[t]
\centering
\resizebox{0.95\linewidth}{!}{%
\begin{tabular}{l|ccc|ccc|ccc}
\hline
\multirow{2}{*}{Model} & \multicolumn{3}{c|}{MUG}                                                                   & \multicolumn{3}{c|}{MHAD}                                                                      & \multicolumn{3}{c}{NATOPS}                                                                     \\ \cline{2-10} 
                       & \multicolumn{1}{c|}{FVD$\downarrow$}    & \multicolumn{1}{c|}{cFVD$\downarrow$}               & sFVD$\downarrow$               & \multicolumn{1}{c|}{FVD$\downarrow$}    & \multicolumn{1}{c|}{cFVD$\downarrow$}                 & sFVD$\downarrow$                 & \multicolumn{1}{c|}{FVD$\downarrow$}    & \multicolumn{1}{c|}{cFVD$\downarrow$}                 & sFVD$\downarrow$                 \\ \hline
ImaGINator \cite{wang2020imaginator}            & \multicolumn{1}{c|}{170.73} & \multicolumn{1}{c|}{257.46\textpm62.88} & 319.37\textpm95.23 & \multicolumn{1}{c|}{889.48} & \multicolumn{1}{c|}{1406.56\textpm260.70} & 1175.74\textpm327.99 & \multicolumn{1}{c|}{721.17} & \multicolumn{1}{c|}{1122.13\textpm150.74} & 1042.69\textpm416.16 \\
VDM \cite{ho2022video}                  & \multicolumn{1}{c|}{108.02} & \multicolumn{1}{c|}{182.90\textpm69.56} & 213.59\textpm97.70 & \multicolumn{1}{c|}{295.55} & \multicolumn{1}{c|}{531.20\textpm104.25}  & 398.09\textpm121.16  & \multicolumn{1}{c|}{169.61} & \multicolumn{1}{c|}{410.71\textpm105.97}  & 350.59\textpm125.03  \\
$\text{LDM}_\text{64}$ \cite{rombach2022high}                  & \multicolumn{1}{c|}{123.88} & \multicolumn{1}{c|}{196.49\textpm66.99} & 236.26\textpm76.08 & \multicolumn{1}{c|}{280.26} & \multicolumn{1}{c|}{515.29\textpm125.70}  & 427.03\textpm112.31  & \multicolumn{1}{c|}{251.72} & \multicolumn{1}{c|}{506.40\textpm125.08}  & 491.37\textpm231.85  \\
$\text{LFDM}_\text{64}$ (Ours)                & \multicolumn{1}{c|}{\textbf{27.57}}  & \multicolumn{1}{c|}{\textbf{77.86\textpm20.27}}  & \textbf{108.36\textpm39.60} & \multicolumn{1}{c|}{\textbf{152.48}} & \multicolumn{1}{c|}{\textbf{339.63\textpm52.88}}   & \textbf{242.61\textpm28.50}   & \multicolumn{1}{c|}{\textbf{160.84}} & \multicolumn{1}{c|}{\textbf{376.14\textpm106.13}}  & \textbf{324.45\textpm116.21}  \\ \hline
$\text{LDM}_\text{128}$ \cite{rombach2022high}                 & \multicolumn{1}{c|}{126.28} & \multicolumn{1}{c|}{208.03\textpm64.86} & 241.49\textpm75.18 & \multicolumn{1}{c|}{337.43} & \multicolumn{1}{c|}{594.34\textpm150.31}  & 497.50\textpm110.16  & \multicolumn{1}{c|}{344.81} & \multicolumn{1}{c|}{627.84\textpm169.52}  & 623.13\textpm320.85  \\
$\text{LFDM}_\text{128}$ (Ours)                 & \multicolumn{1}{c|}{\textbf{32.09}}  & \multicolumn{1}{c|}{\textbf{84.52\textpm24.81}}  & \textbf{114.33\textpm42.62} & \multicolumn{1}{c|}{\textbf{214.39}} & \multicolumn{1}{c|}{\textbf{426.10\textpm63.48}}   & \textbf{328.76\textpm34.42}   & \multicolumn{1}{c|}{\textbf{195.17}} & \multicolumn{1}{c|}{\textbf{423.42\textpm117.06}}  & \textbf{369.93\textpm159.26}  \\ \hline
\end{tabular}%
}
\caption{Quantitative comparison among different methods on multiple datasets for cI2V generation. The 64 and 128 in the subscript of LDM and LFDM indicate that the resolution of synthesized video frames are $64\times 64$ and $128\times 128$, respectively.}
\label{tab:sota_cond}
\vspace{-2mm}
\end{table*}

\section{Experiments}
\subsection{Datasets and Metrics}
We conduct comprehensive experiments on the following video datasets.

\textbf{MUG} facial expression dataset \cite{aifanti2010mug} contains 1,009 videos of 52 subjects performing 7 different expressions, including \textit{anger}, \textit{disgust}, \textit{fear}, \textit{happiness}, \textit{neutral}, \textit{sadness}, and \textit{surprise}. We randomly select 26 subjects for training and use the remaining 26 subjects for testing, which results in 465 and 544 videos in training and testing set, respectively.

\textbf{MHAD} human action dataset \cite{chen2015utd} contains 861 videos of 27 actions performed by 8 subjects. 
This dataset consists of multiple kinds of human actions covering sport actions (\eg, \textit{bowling}), hand gestures (\eg, \textit{draw x}), daily activity (\eg, \textit{stand to sit}) and training exercises (\eg, \textit{lunge}). We randomly choose 4 subjects for training (431 videos) and 4 subjects for testing (430 videos).

\textbf{NATOPS} aircraft handling signal dataset \cite{song2011tracking} includes 9,600 videos of 20 subjects performing 24 body-and-hand gestures used for communicating with the U.S. Navy pilots. It contains some common handling signals such as \textit{spread wings} and \textit{stop}. We randomly select 10 subjects for training (4,800 videos) and 10 subjects for testing (4,800 videos).

\textbf{Data Preprocessing.} We resize all the videos to $128\times 128$ resolution for our models. For MHAD and NATOPS datasets, we also crop video frames by removing some part of background using their provided depth maps. Considering most videos in these datasets are short (no more than 80 frames on average), 
instead of uniformly sampling frames, for each training video, we randomly sample 40 frames and sort them by time to generate diverse video clips with fixed length for training the stage-two DM. 

\textbf{Metrics.} Following prior works \cite{ho2022imagen, ho2022video, hu2022make, skorokhodov2022stylegan}, we compute Fr\'echet Video Distance (\textbf{FVD}) \cite{unterthiner2018towards} to evaluate the \textit{visual quality}, \textit{temporal coherence} and \textit{sample diversity} of generated videos.
Similar to Fr\'echet Inception Distance (FID) \cite{heusel2017gans} used for image quality evaluation, FVD first employs a video classification network I3D \cite{carreira2017quo} pretrained on Kinetics-400 dataset \cite{kay2017kinetics} to obtain feature representation of real and synthesized videos. Then it calculates the Fr\'echet distance between the distributions of real and synthesized video features. To measure how well a generated video corresponds to the class condition $y$ (\textit{condition accuracy}) and the given image $x_0$ (\textit{subject relevance}), similar to the conditional FID in \cite{benny2021evaluation}, we design two variants of FVD: class conditional FVD (\textbf{cFVD}) and subject conditional FVD (\textbf{sFVD}). cFVD and sFVD compare the distance between real and synthesized video feature distributions under the same class condition $y$ or the same subject image $x_0$, respectively. We first compute cFVD and sFVD for each condition $y$ and image $x_0$, then report their mean and variance as final results.
In our experiments, we generate 10,000 videos for all the models to accurately estimate the feature distributions.

\subsection{Model and Baseline Implementation}
\label{subsec:implement}
\textbf{Model Implementation.} 
Our proposed LFDM includes four trainable modules: an image encoder $\Phi$, an image decoder $\Omega$, a flow predictor $F$, and a denoising model $\epsilon_\theta$ from DDPM. These modules are general and can have various backbone networks. 
Here we choose the architecture in \cite{johnson2016perceptual} to implement the encoder $\Phi$ and the decoder $\Omega$ due to its simplicity. For encoder $\Phi$, we use a network with 2 down-sampling blocks, thus the spatial size of latent map $H_z\times W_z$ will be $\frac{1}{4}H_x\times \frac{1}{4}W_x$, only $1/16$ of the size of the input frame $x$.
For decoder $\Omega$, we use a network with 6 residual blocks and two up-sampling blocks. The flow predictor $F$ is implemented with MRAA \cite{siarohin2021motion}, which can estimate latent flow $f$ and occlusion map $m$ based on detected object parts. 
Per MRAA \cite{siarohin2021motion}, we also add the equivariance loss to $L_\text{LFAE}$ in Eq.~\ref{eq:fae}. 
When training LFAE, we set the batch size to be 100 and train it for 100 epochs using the Adam optimizer \cite{kingma2014adam}. The initial learning rate is set to be $2\times10^{-4}$ and drops by a decay factor 0.1 at epoch 60 and 90.

For the denoising model $\epsilon_\theta$ in stage-two DM, we adopt the conditional 3D U-Net architecture in \cite{ho2022video}, which includes 4 down-sampling and 4 up-sampling 3D covolutional blocks. The embedding $e$ of the condition $y$ is concatenated with a time step embedding and then added into each residual blocks of $\epsilon_\theta$. The latent map $z_0$ of image $x_0$ is also provided to $\epsilon_\theta$ by the concatenation with the noise $\mathbf{n}$.
When training DM, we set the batch size to be 20 and train 1,200 epochs using the Adam optimizer \cite{kingma2014adam}. The initial learning rate is set to be $2\times10^{-4}$ and drops by a decay factor 0.1 at epoch 800 and 1000.
We employ the cosine noise schedule \cite{nichol2021improved} and use dynamic thresholding \cite{saharia2022photorealistic} with 90 percentile during the sampling process. To enable stochastic generation, we adopt the training process similar to classifier-free guidance \cite{ho2021classifier}, \ie, the condition embedding $e$ is replaced with a null embedding $\emptyset$ for $\epsilon_\theta$ with 10\%  probability. So stochastic generation can be achieved by simply feeding $\emptyset$ instead of $e$ to $\epsilon_\theta$ during inference (\ie, $g=0$ in Eq.~\ref{eq:classifier}). But we still set $g=1$ for conditional generation instead of using the common $g>1$, because the latter will run two DM forward passes, leading to slower sampling speed \cite{ho2021classifier}. 
To simplify coding, the flow sequence $\hat{\mathbf{f}}$ and occlusion map sequence $\hat{\mathbf{m}}$ also include $f_0$ and $m_0$ (\ie, the flow and occlusion map between $x_0$ and itself). So $\hat{\mathbf{f}}$ and $\hat{\mathbf{m}}$ can have the same size as the output video $\hat{\mathbf{x}}^K_0$.
Unless otherwise specified, we apply $T=1000$ step DDPM to sample 40-frame $32\times 32\times 2$ $\hat{\mathbf{f}}$ and $32\times32 \times 1$ $\hat{\mathbf{m}}$ and finally produce 40-frame videos $\hat{\mathbf{x}}$ with $128\times 128$ frame resolution. 

\begin{table}[t]
\centering
\resizebox{0.65\linewidth}{!}{%
\begin{tabular}{l|cc|c}
\hline
\multirow{2}{*}{Model}                                           & \multicolumn{2}{c|}{FVD$\downarrow$} & \multirow{2}{*}{Gap$\downarrow$} \\ \cline{2-3}
                                                                 & \multicolumn{1}{c|}{Train}  & Test   &                                  \\ \hline
ImaGINator \cite{wang2020imaginator}            & \multicolumn{1}{c|}{15.92}  & 170.73 & 154.81                           \\
VDM \cite{ho2022video}                         & \multicolumn{1}{c|}{17.39}  & 108.02 & 90.63                            \\
$\text{LDM}_\text{64}$ \cite{rombach2022high}  & \multicolumn{1}{c|}{55.44}  & 123.88 & 68.44                            \\
$\text{LFDM}_\text{64}$ (Ours)                                  & \multicolumn{1}{c|}{\textbf{10.58}}  & \textbf{27.57}  & \textbf{16.99}  \\ \hline
$\text{LDM}_\text{128}$ \cite{rombach2022high} & \multicolumn{1}{c|}{67.51}  & 126.28 & 58.77                            \\
$\text{LFDM}_\text{128}$ (Ours)                                 & \multicolumn{1}{c|}{\textbf{14.17}}  & \textbf{32.09}  & \textbf{17.92}  \\ \hline
\end{tabular}
}
\caption{Comparison of different models conditioned on image $x_0$ from either the training set or testing set of MUG dataset.}
\label{tab:train_test}
\vspace{-2mm}
\end{table}

\textbf{Baseline Implementation.} 
We compare LFDM with three baseline models, including GAN-based I2V model \textbf{ImaGINator} \cite{wang2020imaginator}, video diffusion models \textbf{VDM} \cite{ho2022video}, and a variant of image latent diffusion models \textbf{LDM} \cite{wang2022latent}. We extend LDM to the video domain by replacing the 2D U-Net in their DM with 3D U-Net.
For a fair comparison, VDM, LDM and our LFDM use the same 3D U-Net architecture and the same way to utilize conditioning information by concatenation. 
LDM is designed to have the same-size latent space as LFDM ($40\times 32\times 32\times 3$) and also generate 40-frame videos with $128\times 128$ resolution. Due to high computational cost, we only use VDM to generate 40-frame videos with $64\times 64$ resolution.
ImaGINator is implemented with their provided code, which generates 32-frame videos with $64\times 64$ resolution. For a fair comparison, when computing metrics, all the real videos are resized to be the same volume size as the generated videos for each model. We also report the results of LDM and LFDM under the $64\times64$ resolution. For sampling, we use 1000-step DDPM for LDM and LFDM. Since DDPM sampling is very slow in the large latent space of VDM ($40\times64\times64\times3$), we employ 200-step DDIM \cite{song2020denoising} to accelerate the sampling process. We find that this method can achieve comparable performance as DDPM for VDM in our preliminary experiments.

\subsection{Result Analysis}
\label{subsec:results}
\textbf{Conditional Generation.}
Table~\ref{tab:sota_cond} shows the quantitative comparison between our method and baseline models for conditional image-to-video (cI2V) task, where our proposed LFDM consistently outperforms all the other baseline models under the resolution of either $64\times 64$ or $128\times 128$. As Fig.~\ref{fig:sota} shows, ImaGINator and LDM both fail to produce fine-grained details while VDM sometimes just generates some almost-still frame sequences (\eg, results on MHAD in Fig~\ref{fig:sota}). 
ImaGINator may suffer from the limited synthesis ability of their used GAN model, while LDM and VDM may sometimes have difficulty in modeling their latent spaces, which couple spatial and temporal features.
Though VDM achieves similar FVD scores to ours on NATOPS dataset, note that VDM has higher computational cost than LFDM since the latent flow space in our DM has lower dimensions. 
When sampling with the batch size of 10 using 1000-step DDPM on one NVIDIA A100 GPU, LFDM costs about 0.9GB and 36s to generate one video of $128\times 128$ resolution while VDM requires about 2.5GB and 112.5s to sample one video of only $64\times 64$ resolution.

\begin{table}[t]
\centering
\resizebox{0.7\linewidth}{!}{%
\begin{tabular}{l|ccc}
\hline
\multirow{2}{*}{Model} & \multicolumn{3}{c}{$L_1$ error$\downarrow$}                                             \\ \cline{2-4} 
                       & \multicolumn{1}{c|}{MUG}    & \multicolumn{1}{c|}{MHAD}   & NATOPS \\ \hline
LDM \cite{rombach2022high}                   & \multicolumn{1}{c|}{0.419} & \multicolumn{1}{c|}{\textbf{0.294}} & \textbf{0.222} \\
LFDM (Ours)                   & \multicolumn{1}{c|}{\textbf{0.418}} & \multicolumn{1}{c|}{0.302} & 0.260 \\ \hline
\end{tabular}%
}
\caption{Comparison of self-reconstruction $L_1$ error between the auto-encoder of LDM and that of LFDM on testing sets of multiple datasets under the resolution of $128\times128$. }
\label{tab:self-rec}
\vspace{-2mm}
\end{table}

\begin{table}[t]
\centering
\resizebox{0.65\linewidth}{!}{%
\begin{tabular}{l|c|c}
\hline
Model      & FVD$\downarrow$    & sFVD$\downarrow$               \\ \hline
ImaGINator \cite{wang2020imaginator} & 196.18 & 352.52\textpm104.54 \\
VDM \cite{ho2022video}       & 106.66 & 214.21\textpm95.96  \\
$\text{LDM}_\text{64}$ \cite{rombach2022high}       & 125.36 & 242.19\textpm79.40  \\
$\text{LFDM}_\text{64}$ (Ours)      & \textbf{42.50}  & \textbf{154.50\textpm50.54}  \\ \hline
$\text{LDM}_\text{128}$ \cite{rombach2022high}        & 127.12 & 247.28\textpm74.26  \\
$\text{LFDM}_\text{128}$ (Ours)      & \textbf{49.31}  & \textbf{167.58\textpm56.85}  \\ \hline
\end{tabular}%
}
\caption{Comparison among different methods on MUG dataset for stochastic image-to-video generation.}
\label{tab:uncond}
\vspace{-2mm}
\end{table}

To further analyze the performance difference among different models, we also compute their FVD scores of generated videos conditioned on the image $x_0$ from the training set of MUG dataset. As Table~\ref{tab:train_test} shows, all three baseline models have much better performance when conditioned on training (\textit{seen}) images than testing (\textit{unseen}) images, while our proposed LFDM noticeably shows the least training-testing gap. This can be attributed to our warp-based design, two-stage disentangled training framework, among others. We also notice that LDM fails to achieve low FVD scores like other methods when conditioned on training images. Considering that the auto-encoder in LDM has already achieved similar or even better reconstruction performance than LFDM (see Table~\ref{tab:self-rec}), we speculate that the 3D U-Net in LDM may not be effective enough to model the latent space that encodes both spatial and temporal features. However, built on the same 3D U-Net architecture, LFDM can successfully synthesize visually-appealing and temporally-coherent videos, illustrating simpler latent space in our DM, which only contains motion and shape information.
From Table~\ref{tab:sota_cond}, one can also observe that all the models achieve better FVD results on MUG dataset than on MHAD and NATOPS datasets. The reason may be that facial videos contain less high-frequency spatio-temporal details than human movement videos.

\textbf{Stochastic Generation.}
As mentioned in Sec.~\ref{subsec:implement}, by replacing the condition embedding $e$ with the null embedding $\emptyset$ during inference, we enable VDM, LDM, and our LFDM for stochastic generation that only depends on image $x_0$. We also retrain ImaGINator for stochastic generation by removing the condition input $y$. As Table~\ref{tab:uncond} shows, our proposed LFDM still outperforms all the other baseline models on MUG dataset, validating the effectiveness of LFDM on both conditional and stochastic I2V generation.

\begin{figure}
    \centering
    \includegraphics[width=0.75\linewidth]{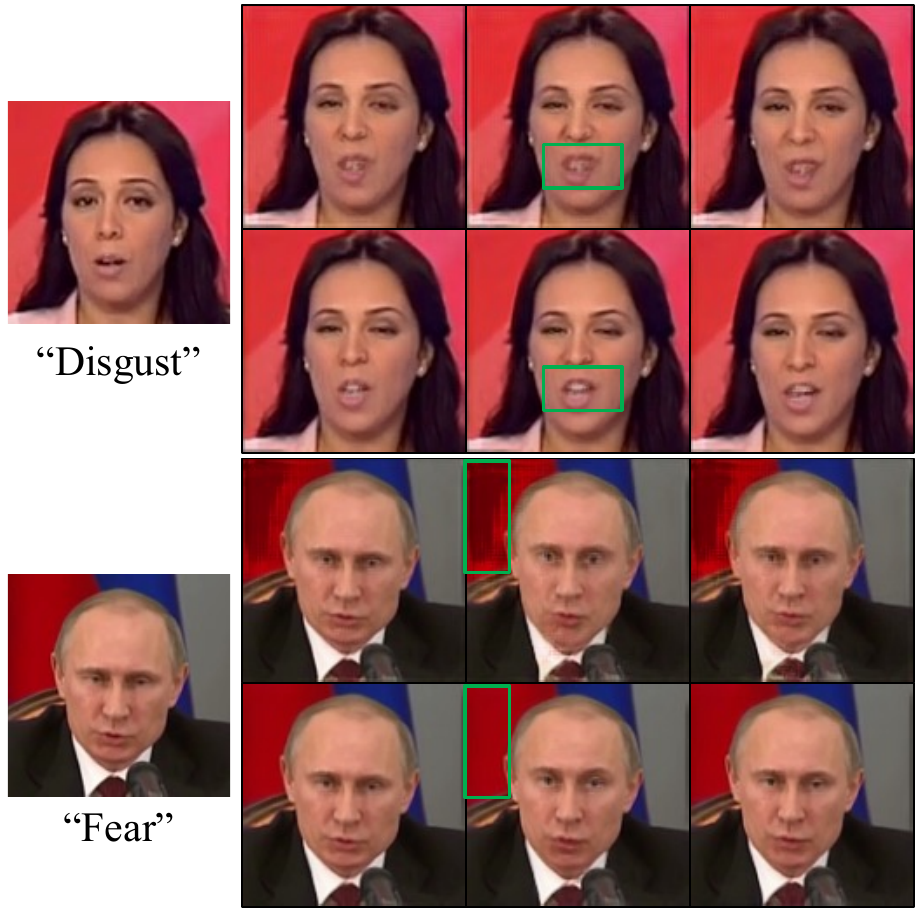}
    \caption{Qualitative comparison of LFDM with original (the 1st\&3rd rows) \vs finetuned (the 2nd\&4th rows) decoder on FaceForensics dataset \cite{rossler2018faceforensics}. The first column shows the given image $x_0$ and condition $y$. The green boxes highlight differences.}
    \label{fig:newdomain}
    \vspace{-2mm}
\end{figure}

\textbf{Application to New-domain Facial Videos.}
We also apply our proposed LFDM trained on MUG dataset to new-domain facial videos. Here we choose \textbf{FaceForensics} \cite{rossler2018faceforensics} dataset, which contains 1,004 subject videos of news briefing from different reporters. We randomly choose 150 subjects for training and 150 subjects for testing, and we utilize a face alignment algorithm \cite{king2009dlib} to extract the facial regions of video frames. 
We compare two models: (1) directly applying the original trained LFDM, and (2) finetuning the decoder $\Omega$ with the unsupervised training loss $L_\text{LFAE}$ in Eq.~\ref{eq:fae} using the training videos from FaceForensics dataset, but freezing all the remaining modules. 
Due to the lack of ground truth videos, to measure performance, we instead calculate image-based \textbf{FID} \cite{heusel2017gans} and subject FID (\textbf{sFID}) scores. Similar to sFVD, sFID first computes the FID scores between distributions of real and synthesized frames for each subject image $x_0$ and then reports their mean and variance. We also report the $L_1$ error on testing set to compare the self-reconstruction performance of LFAE using original and finetuned $\Omega$. As Table~\ref{tab:newdomain} and Fig.~\ref{fig:newdomain} show, by simply finetuning the decoder $\Omega$ with \textit{unlabeled} new videos, LFDM can still achieve promising performance on new-domain facial videos. 
This illustrates the flexibility of our two-stage training framework.
To improve spatial content quality, one can just finetune the decoder in stage-one LFAE, without the need to retrain the whole framework. 

\textbf{Ablation Study.} We conduct ablation study of different sampling strategies in Table~\ref{tab:aba_sampling}.
For each setting, we report FVD scores and the average sampling time to generate one video when using the batch size of 10 on one NVIDIA A100 GPU. 
We first compare the results sampled by 10-step DDIM and 100-step DDIM against 1000-step DDPM (default setting). 
It is interesting to observe that 10-step DDIM shows noticeably better performance than 100-step DDIM with faster sampling speed. 
We also try two typical values of guidance scale $g$ in Eq.~\ref{eq:classifier}, small guidance $1.1$ and large guidance $5.0$.  
Though using $g=1.1$ guidance slightly improves the FVD score, it also doubles the inference time. So we use $g=1$ as our default setting.
We also try to change the network depth of the decoder $\Omega$ in stage-one LFAE and the denoising model $\epsilon_\theta$ in stage-two DM; results of this ablation study, generated video examples and further discussion are included in our Supp. materials.

\begin{table}[t]
\centering
\resizebox{0.9\linewidth}{!}{%
\begin{tabular}{l|c|c|c}
\hline
Model                      & $L_1$ error$\downarrow$ & FID$\downarrow$   & sFID$\downarrow$              \\ \hline
Original decoder $\Omega$  & 2.129      & 38.48 & 75.08\textpm37.33 \\
Finetuned decoder $\Omega$ & \textbf{1.310}      & \textbf{23.36} & \textbf{51.56\textpm26.36} \\ \hline
\end{tabular}%
}
\caption{Quantitative comparison of LFDM with original \vs finetuned decoder on FaceForensics dataset.}
\label{tab:newdomain}
\vspace{-2mm}
\end{table}

\begin{table}[t]
\centering
\resizebox{0.65\linewidth}{!}{%
\begin{tabular}{l|c|c|c}
\hline
Steps     &$g$-scale &  FVD$\downarrow$    & Time(s)$\downarrow$  \\ \hline
DDIM-10   & $1.0$     & 50.18  & \textbf{0.3}      \\
DDIM-100  & $1.0$     & 106.11 & 3.5      \\ \hline
DDPM-1000 & $1.1$     & \textbf{31.84}  & 71.4     \\
DDPM-1000 & $5.0$     & 49.82  & 71.4     \\ \hline
DDPM-1000 & $1.0$     & {32.09}  & 36.0     \\ \hline
\end{tabular}%
}
\caption{Ablation study comparing different sampling strategies for LFDM on MUG dataset (resolution is $128\times128$). }
\label{tab:aba_sampling}
\vspace{-2mm}
\end{table}

\section{Conclusion and Discussion}
In this paper, we propose a novel conditional image-to-video framework, latent flow diffusion models (LFDM), to generate videos by warping given images with flow sequences generated in the latent space based on class conditions. 
Comprehensive experiments show that LFDM can achieve state-of-the-art performance on multiple datasets.

Though achieving promising performance, our proposed LFDM still suffers from several limitations. 
First, current experiments with LFDM are limited to videos containing a single moving subject. 
We plan to extend the application of LFDM to multi-subject flow generation in the future.
Second, the current LFDM is conditioned on class labels instead of natural text descriptions. 
Text-to-flow is an interesting topic and we leave this direction as future work. 
Finally, compared with GAN models, LFDM is much slower when sampling with 1000-step DDPM. In the future, we plan to further investigate some fast sampling methods \cite{lu2022dpm,kong2021fast} in order to reduce generation time.
{\small
\bibliographystyle{ieee_fullname}
\bibliography{egbib}
}

\end{document}


\title{Supplementary Materials for CVPR'23 Paper Titled \\
``Conditional Image-to-Video Generation with Latent Flow Diffusion Models''}

\maketitle


\section{Potential Negative Social Impact} Conditional image-to-video models can be used for unethical purposes \cite{yu2022generating}, \eg, creating videos of celebrities for fake news spreading. We will restrict the usage of our models to research purposes only. We also plan to investigate some fake video detection techniques \cite{amerini2019deepfake} that may be effective in detecting fake videos like the ones generated by our methods.

\section{Additional Experiments}
\subsection{Additional Ablation Study on Network Architecture}

To evaluate the performance difference of our proposed LFDM with different architectures, we change the depth of the image decoder $\Omega$ in stage-one LFAE (Table~\ref{tab:aba_decoder}) and the 3D U-Net $\epsilon_\theta$ in stage-two DM (Table~\ref{tab:aba_dm}). We experiment with different settings on MUG dataset to generate videos of $128\times128$ frame resolution. 

In our default setting, the image decoder $\Omega$ in stage-one LFAE is implemented with a network including 6 residual blocks and 2 up-sampling blocks. In Table~\ref{tab:aba_decoder}, we compare using different network depths for the image decoder $\Omega$ in stage-one LFAE . We add four extra residual blocks to the decoder $\Omega$. So the number of residual blocks is increased from 6 to 10. Then we only retrain this deeper decoder in stage one, while keeping all the remaining modules unchanged. As Table~\ref{tab:aba_decoder} shows, using a deeper image decoder shows slightly better self-reconstruction performance (as measured by $L_1$ error) but fails to generate higher-quality videos (as measured by FVD).
Therefore, we keep using 6 residual blocks in our experiments.

In our default setting, the denoising network $\epsilon_\theta$ employs a 3D U-Net architecture including 4 down-sampling and 4 up-sampling 3D convolutional blocks, where the \textit{channel multipliers} are (1, 2, 4, 8) with a base channel of 64. That is, from highest to lowest resolution,  the 4 down- or up-sampling blocks in $\epsilon_\theta$ use ($1\times64$, $2\times64$, $4\times64$, $8\times64$) channels, respectively. In Table~\ref{tab:aba_dm}, we compare using different channel multipliers in stage-two DM . We add one more layer to the down-sampling and up-sampling blocks of the 3D U-Net and the {channel multipliers} are (1, 2, 4, 8, 16) with a base channel of 64. We retrain this deeper DM in stage two with 1,200 training epochs as in our previous simpler DM training. We keep using the same stage-one LFAE. From Table~\ref{tab:aba_dm}, one can observe that using more layers in DM led to decreased performance. 
Therefore, we adopt the simpler (1, 2, 4, 8) as the default setting of channel multipliers in our stage-two DM.

\begin{table}[t]
    \centering
\resizebox{0.6\linewidth}{!}{%
\begin{tabular}{c|c|c}
\hline
\# Residual Blocks  & $L_1$ error$\downarrow$ & FVD$\downarrow$   \\ \hline
6  & 0.418       & \textbf{32.09} \\
10 & \textbf{0.371}       & 32.83     \\ \hline
\end{tabular}%
}
\caption{Comparison using different numbers of residual blocks in the image decoder $\Omega$ of stage-one LFAE.}
\label{tab:aba_decoder}
\end{table}

\begin{table}[t]
    \centering
    \resizebox{0.5\linewidth}{!}{%
\begin{tabular}{c|c}
\hline
Channel Multipliers & FVD$\downarrow$   \\ \hline
(1, 2, 4, 8)     & \textbf{32.09} \\
(1, 2, 4, 8, 16) & 68.07 \\ \hline
\end{tabular}%
}
\caption{Comparison using different channel multipliers in the network $\epsilon_\theta$ of stage-two DM.}
\label{tab:aba_dm}
\end{table}

\subsection{Additional Analysis of Flow and Occlusion Maps}
\begin{figure}[h]
    \centering
\includegraphics[width=\linewidth]{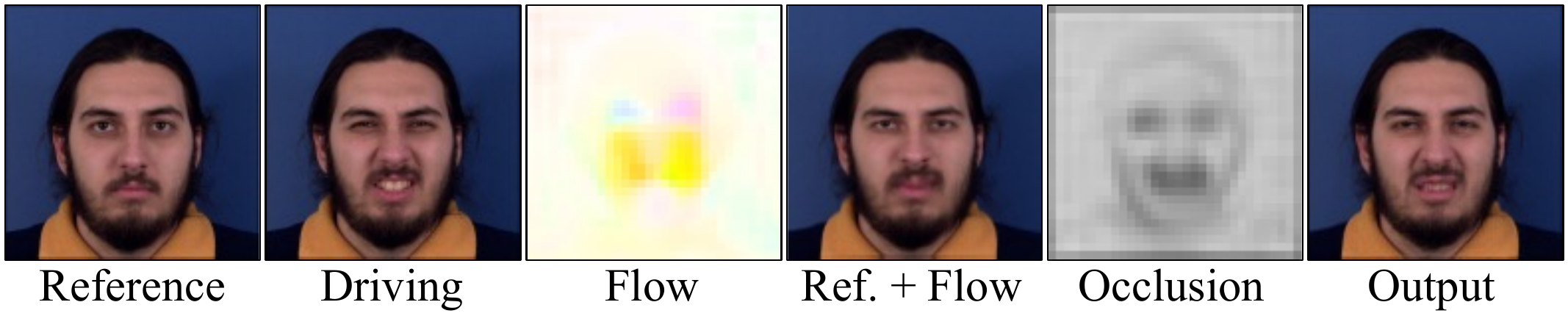}
    \caption{Visualization of flow and occlusion map. [Ref. + Flow] is generated by applying the flow to reference image; note the change in head pose and the shape of eyes and mouth after applying flow. Occlusion map further masks the eyes and mouth to help decoder generate novel pixels for these parts in output image. }
    \label{fig:vis}
\end{figure}
Figure~\ref{fig:vis} shows the visualization of flow and occlusion map of one example video frame from MUG dataset. As illustrated in the caption of Fig.~\ref{fig:vis}, without using occlusion map, decoder may need to learn which regions should be kept and which regions should be masked and repainted. Our additional experiments show that retraining LFAE without occlusion maps increases the $L_1$ error of self-reconstruction from 0.418 to 0.450 on MUG dataset. 

\subsection{Comparison of Inference Time among Different Models}
Table~\ref{tab:time} shows the average inference time of each method to generate one video when using batch size 10 on one NVIDIA A100 GPU on MUG dataset. Note that VDM uses 200-step DDIM while both LDM and LFDM employ 1000-step DDPM.
\begin{table}[h]
\resizebox{\linewidth}{!}
{%
\begin{tabular}{l|cccc|cc}
\hline
Model    & ImaGINator & VDM  & LDM$_\text{64}$ & LFDM$_\text{64}$ & LDM$_\text{128}$  & LFDM$_\text{128}$ \\
\hline
Time(s) & 0.9        & 23.1 & 8.0 & 8.8  & 25.5 & 36.0 \\
\hline
\end{tabular}%
}
\caption{Inference time comparison among different methods.}
\label{tab:time}
\end{table}

\section{More Discussion about Future Work}
Several limitations and some future work are discussed in Section~5 of the paper. Here we elaborate more on future work about LFDM. One future direction is to enable the generation of a video with changing background (or context). We plan to first utilize our LFDM to generate a video describing the motion of foreground subject, and then design another generative network conditioned on each generated foreground frame to synthesize the changing background for each frame. 
In addition, to enhance the generalization ability of LFDM on generating diverse motions of more categories, 
we plan to collect more labeled training video datasets and apply some continual/incremental learning techniques such as  \cite{volpi2021continual,wang2022learning,liang2022balancing,wang2022dualprompt} to train our LFDM. 
Finally, in our experiments (Table~6), we noticed that 10-step DDIM can achieve acceptable generation performance with faster sampling speed, suggesting it may have greater potential with better hyperparameter settings. To explore these settings, including diffusion sampling steps, we plan to employ some recent hyperparameter optimization techniques such as \cite{bergstra2012random,li2017hyperband,zhang2023targeted}. 

\section{Information about Attached Videos}
We attach seven MP4 files of example video clips generated by our proposed method in Supp. materials\footnote{These videos are also available in \url{https://github.com/nihaomiao/CVPR23_LFDM}.}. All the given images are testing (\textit{unseen}) images. 
\begin{itemize}
    \item \textbf{mug.mp4} shows the synthesized video clips displaying all 7 expressions of one subject from MUG dataset.
    \item \textbf{mhad1.mp4} and \textbf{mhad2.mp4} include the generated video clips for 26 actions of one subject from MHAD dataset. We exclude the action \textit{sit to stand} because the subject in the given image is standing.
    \item \textbf{natops.mp4} shows the synthesized video clips containing all 24 gestures of one subject from NATOPS dataset.
    \item \textbf{new\_domain.mp4} shows the synthesized video clips including 4 expressions of four subjects from FaceForensics dataset. ``Original'' means directly applying our LFDM pretrained on MUG dataset. ``Finetuned'' means that the image decoder is finetuned with the \textit{training} videos from FaceForensics dataset. Note that other modules including stage-two DM are still unchanged during finetuning. From this video, one can observe that our original LFDM can generate acceptable results for given subject images from a new domain and achieve better performance when the decoder is finetuned with training videos from the new domain. 
    \item \textbf{mug\_ddim.mp4} shows the synthesized video clips containing 4 expressions of four subjects from MUG dataset.
    ``DDIM-10'' means using 10-step DDIM for diffusion sampling while ``DDPM-1000'' is our default 1000-step DDPM sampling strategy. From this video, one can observe that 10-step DDIM can generate visually-acceptable videos with faster sampling speed (0.3s per video \vs 36s per video when using DDPM-1000). But note that the FVD score  of DDPM-1000 is still noticeably better than DDIM-10 (32.09 \vs 50.18) so we keep DDPM-1000 as our default setting.
    
    \item \textbf{sota.mp4} is a video for comparison between our proposed LFDM and several other models including ImaGINator, VDM, and LDM.  We show synthesized video clips by each model on 3 subjects from MUG, MHAD, and NATOPS datasets. The video frames of ground truth (GT) and results of LDM and our LFDM have $128\times128$ resolution while results of ImaGINator and VDM are $64\times64$. The original video clips generated by ImaGINator only contain 32 frames. So we repeat the first frame and the last frame four times to make all the displaying videos have 40 frames. 
\end{itemize}

{\small
\bibliographystyle{ieee_fullname}
\bibliography{egbib}
}